# From Relational Databases to Belief Networks

Wilson X. Wen
AI Systems, Telecom Research Labs.
770 Blackburn Rd, Clayton,
Victoria 3168, Australia

## Abstract

The relationship between belief networks and relational databases is examined. Based on this analysis, a method to construct belief networks automatically from statistical relational data is proposed. A comparison between our method and other methods shows that our method has several advantages when generalization or prediction is deeded.

## 1 INTRODUCTION

It turns out that Relational Database (RD) and Belief Network (BN) are very closely related to each other in many aspects. Spiegelhalter [1986] investigated some of these relationships. He also discussed the issue about "using data to learn about quantitative assessments" so that the conditional probabilities can be revised by data obtained after a BN has been built. Lauritzen, Spiegelhalter [1988], Herskovitz and Copper [1990] used a method based on Maximum Entropy (ME) principle [Shore and Johnson, Jan 1980] to obtain a consistent distribution from empirical data. Spiegelhalter [1986] and Wen [1990b] discussed decomposition of the networks to reduce the computational amount required by probabilistic reasoning.

In this paper, the relationship between RD and BN is investigated and a method to construct BN from statistical RD [Wen, 1990a] is proposed based on the principles of Nearest Neighborhood (NN) [Duda and Hart, 1973] and Occam's Razor (OR) [Blumer et al., 1987]. Most of the contemporary databases are relational, this makes the research in construction of BN from RD interesting and important. A comparison between our method and others is also given.

## 2 RELATIONAL DATABASES

According to the relational database theory [Ullman, 1982], we have the following basic definitions in RD:

A *relation r* is a subset of the Cartesian product of domains $D_1, ..., D_k$. A *domain $D_i$* is a set of values taken by an *attribute $A_i$*. The members of a relation are *tuples*. The value of a tuple $t$ on attribute $A$ is written as $t[A]$. The set of attribute for a relation $r$ is the *relation scheme R*. Let $X \subset R$ and $t \in r$. We write $t[X]$ for the partial tuple of $t$ restricted to $X$. A collection of relation schemes is a *relational database scheme*. The current values of the relations corresponding to the database scheme are the *relational database*.

Let $r$ and $s$ be relations on relational schemes $R$ and $S$, respectively, $A \in R$, $a \in D$, and $X \subset R$. We will discuss the following operators on relations.

1. **Select operator:** "Select from $r$ with $A$ equal to $a$" yields a relation $\sigma_{A=a}(r) = \{t \in r | t[A] = a\}$.

2. **Project operator:** The projection of $r$ onto $X$ is a relation $\pi_X(r) = \{t[X] | t \in r\}$

3. **Join operator:** The join of $r$ and $s$ is a relation $r \bowtie s = \{t | t[R] \in r \wedge t[S] \in s\}$

To avoid redundancy and potential inconsistency, RD are often organized in normal forms according to the dependencies existing among the attribute subsets.

**Functional dependency (FD):** Let $X, Y \subset R$. $X$ functionally determines $Y$, written $X \to Y$, if

$$\forall r \text{ on } R, \ t, s \in r \wedge t[X] = s[X] \Longrightarrow t[Y] = s[Y].$$

**Multivalued dependency (MD):** $X$ multidetermines $Y$, written $X \to\!\!\!\to Y$, if $\forall r$ on $R$, $t, s \in r \wedge t[X] = s[X]$ implies that $\exists u, v \in r$ such that (1) $u[X] = v[X] = t[X] = s[X]$, (2) $u[Y] = t[Y]$, and $u[R - X - Y] = s[R - X - Y]$, and (3) $v[Y] = s[Y]$ and $v[R - X - Y] = t[R - X - Y]$.

**Join dependency (JD)** over $R_1, ..., R_n$, written $\bowtie (R_1, ..., R_n)$, is satisfied by a relation $r$ over $R_1 \cup ... \cup R_n$, if and only if $\pi_{R_1}(r) \bowtie ... \bowtie \pi_{R_n}(r) = r$

Let $F$ be a set of FD's on $R(A_1, ..., A_n)$ and $F^+$ the closure of $F$. $K \subset R$ is a *key* of $R$ if $K \to A_1, ..., A_n \in F^+ \wedge \not\exists X \subset K, X \to A_1, ..., A_n \in F^+$.

To determine keys and calculate $F^+$, a set of inference rules, which is both complete and sound, called Armstrong's axioms has been developed [Ullman, 1982]. According to these rules $R(A_1, ..., A_n)$ can be decomposed into a collection of subsets $\rho(R_1, ..., R_k)$ such that $R = R_1 \cup, ... \cup R_k$. For a decomposition $\rho$ of $R$ the following properties are always desirable:

**Lossless Join (LJ):** Suppose $D$ is a set of dependencies in $R$. $\rho$ has a lossless join w.r.t. $D$ if $\forall r$ on $R, r = \pi_{R_1}(r) \bowtie ... \bowtie \pi_{R_k}(r)$. With this property any relation can be recovered from its projections.

**Dependency Preservation (DP):** This requires that $D$ is preserved by the projection $\pi_{R_i}(D)$ of $D$ onto $R_i$'s, where $\pi_{R_i}(D) = \{< X, Y > \mid < X, Y > \in D^+\}$, $XY \subseteq R_i$, $XY = X \cup Y$ and $< X, Y >$ represents either $X \to Y$ or $X \to\to Y$. $\rho$ preserves $D$ if

$$\forall < X, Y > \in D, < X, Y > \in \bar{D}^+, \quad \bar{D} = \bigcup_{i=1}^{k} \pi_{R_i}(D).$$

The normal forms we are going to discuss include:

**Fourth Normal Form (4NF):** $R$ is in 4th normal form if $\forall X \to\to Y \in D$,

$$Y \neq \emptyset \land Y \not\subseteq X \land XY \neq R \Longrightarrow \exists \text{ a key } K, \ K \subseteq X.$$

**Acyclic Databases:** The relations of the database form an acyclic hypergraph [Beeri et al., 1983].

The following algorithms [Ullman, 1982] decomposes $R$ into a 4NF decomposition with LJ and DP.

**Algorithm 2.1:**

**Input:** Relation scheme $R$ and set of FD and MD $D = \{< X, A >\}$.

**Output:** A 4NF decomposition $\rho$ of $R$ with LJ and DP.

**Method:** There are three cases to be discussed:

1. If $\exists A' \in R, \forall < X, A > \in D, A \neq A' \land A' \notin X$ then $\rho$ contains a relation scheme with only one element $A'$.

2. If $\exists < X, A > \in D, \ XA = R$ then the output decomposition is $R$ itself.

3. Otherwise, $\rho$ contains scheme $XA$ for each $< X, A > \in D$.

Finally, $\rho$ should also contain a relation scheme $K$ which is a key of the original relation $R$. $\square$

**Example 1. Sarcophagal Disease**

The model [Gallant, 1988] consists of 6 symptoms, 2 diseases and 3 possible treatments in Table 1. The three columns in Table 1 forms a simple database. All variables are binary variables taking values 1 and -1, except variables $u_3$ and $u_5$ which can take values 1, 0, and -1. Here 1 means that the corresponding proposition is true, -1 means false, and 0 means unknown. Suppose that there is a set $F$ of FD on $R(u_1, ..., u_{11})$:

$$u_1, u_2, u_3 \quad \to \quad u_7. \quad u_3, u_4, u_5 \quad \to \quad u_8.$$
$$u_6, u_7, u_8 \quad \to \quad u_9. \quad u_3, u_7, u_8 \quad \to \quad u_{10}.$$
$$u_9, u_{10} \quad \to \quad u_{11}.$$

| Symptom | | | | | | Disease | | Treatment | | |
|---|---|---|---|---|---|---|---|---|---|---|
| $u_1$ | $u_2$ | $u_3$ | $u_4$ | $u_5$ | $u_6$ | $u_7$ | $u_8$ | $u_9$ | $u_{10}$ | $u_{11}$ |
| 1 | 1 | 1 | -1 | 0 | -1 | 1 | -1 | 1 | -1 | -1 |
| -1 | -1 | -1 | 1 | 1 | -1 | -1 | 1 | 1 | 1 | -1 |
| -1 | -1 | 1 | 1 | -1 | 1 | 1 | 1 | -1 | -1 | -1 |
| 1 | 1 | -1 | -1 | 1 | -1 | -1 | -1 | -1 | -1 | -1 |
| 1 | -1 | 0 | 1 | 1 | 1 | 1 | 1 | -1 | 1 | 1 |
| 1 | -1 | -1 | 1 | 1 | -1 | 1 | 1 | 1 | 1 | -1 |
| 1 | 1 | 1 | 1 | -1 | 1 | 1 | -1 | -1 | -1 | -1 |
| -1 | 1 | 1 | -1 | 1 | 1 | -1 | 1 | -1 | -1 | -1 |

| | | |
|---|---|---|
| $u_1$:Swollen feet | $u_2$:Read ears | $u_3$:Hair loss |
| $u_4$:Dizziness | $u_5$:Sensitive aretha | |
| $u_6$:Placibin allergy | $u_7$:Supercilliosis | $u_8$:Namastosis |
| $u_9$ :Placibin | $u_{10}$:Biramibio | $u_{11}$:Posiboost |

Table 1: Data Set of Sarcophagal Disease

According to Algorithm 2.1, we have a decomposition of $R$ with a lossless join and preservation of dependencies in $F$ which contains the following relation schemes:

$$R_1 = \{u_1, u_2, u_3, u_7\}. \quad R_2 = \{u_3, u_4, u_5, u_8\}.$$
$$R_3 = \{u_6, u_7, u_8, u_9\}. \quad R_4 = \{u_3, u_7, u_8, u_{10}\}.$$
$$R_5 = \{u_9, u_{10}, u_{11}\}. \quad R_6 = \{u_1, u_2, u_3, u_4, u_5, u_6\}.$$

where $R_6$ is a key of $R$. Thus, we have a relational sample database in 4th normal form in Table 2. $\square$

| $u_1$ | $u_2$ | $u_3$ | $u_7$ | $u_3$ | $u_4$ | $u_5$ | $u_8$ |
|---|---|---|---|---|---|---|---|
| 1 | 1 | 1 | 1 | 1 | -1 | 0 | -1 |
| -1 | -1 | -1 | -1 | -1 | 1 | 1 | 1 |
| -1 | -1 | 1 | 1 | 1 | 1 | -1 | 1 |
| 1 | 1 | -1 | -1 | -1 | -1 | 1 | -1 |
| 1 | -1 | 0 | 1 | 0 | 1 | 1 | 1 |
| 1 | -1 | -1 | 1 | -1 | 1 | 1 | 1 |
| 1 | 1 | 1 | 1 | 1 | -1 | -1 | -1 |
| -1 | 1 | 1 | -1 | 1 | -1 | 1 | 1 |

| $u_6$ | $u_7$ | $u_8$ | $u_9$ | $u_3$ | $u_7$ | $u_8$ | $u_{10}$ |
|---|---|---|---|---|---|---|---|
| -1 | 1 | -1 | 1 | 1 | 1 | -1 | -1 |
| -1 | -1 | 1 | 1 | -1 | -1 | 1 | 1 |
| 1 | 1 | 1 | -1 | 1 | 1 | 1 | -1 |
| -1 | -1 | -1 | -1 | -1 | -1 | -1 | -1 |
| 1 | 1 | 1 | -1 | 0 | 1 | 1 | 1 |
| -1 | 1 | 1 | 1 | -1 | 1 | 1 | 1 |
| 1 | 1 | -1 | -1 | 1 | 1 | -1 | -1 |
| 1 | -1 | 1 | -1 | 1 | -1 | 1 | -1 |

| $u_9$ | $u_{10}$ | $u_{11}$ | $u_1$ | $u_2$ | $u_3$ | $u_4$ | $u_5$ | $u_6$ |
|---|---|---|---|---|---|---|---|---|
| 1 | -1 | -1 | 1 | 1 | 1 | -1 | 0 | -1 |
| 1 | 1 | -1 | -1 | -1 | -1 | 1 | 1 | -1 |
| -1 | -1 | -1 | -1 | -1 | 1 | 1 | -1 | 1 |
| -1 | -1 | -1 | 1 | 1 | -1 | -1 | 1 | -1 |
| -1 | 1 | 1 | 1 | -1 | 0 | 1 | 1 | 1 |
| 1 | 1 | -1 | 1 | -1 | -1 | 1 | 1 | -1 |
| -1 | -1 | -1 | 1 | 1 | 1 | 1 | -1 | 1 |
| -1 | -1 | -1 | -1 | 1 | 1 | -1 | 1 | 1 |

Table 2: A sample database in 4th normal form

**Example 2 [Cooper, 1984]:** Metastatic cancer (A) is a possible cause of a brain tumor (C) and is also an explanation for increased total serum calcium (B). In turn, either of these could explain a patient falling into a coma (D). Severe headache (E) is also possibly associated with a brain tumor.

Suppose we have the statistical information in Table 3 from a sample database. Each entry of the table gives the number of occurrences of the records in the database which contain various combinations of the attributes A, B, C, D, and E. This can be obtained by a projection of the universal relation of the database on the set $ABCDE$ without eliminating the duplicates. Obviously, the following MD hold on ABCDE:

$$A \longrightarrow\!\!\!\rightarrow B, \ A \longrightarrow\!\!\!\rightarrow C, \ B,C \longrightarrow\!\!\!\rightarrow D, \ C \longrightarrow\!\!\!\rightarrow E.$$

|  | $\bar{D}\bar{E}$ | $\bar{D}E$ | $D\bar{E}$ | $DE$ |
|---|---|---|---|---|
| $\bar{A}\bar{B}\bar{C}$ | 23104 | 34656 | 1216 | 1824 |
| $\bar{A}\bar{B}C$ | 128 | 512 | 512 | 2048 |
| $\bar{A}B\bar{C}$ | 1216 | 1824 | 4864 | 7296 |
| $\bar{A}BC$ | 32 | 128 | 128 | 512 |
| $A\bar{B}\bar{C}$ | 1216 | 1824 | 64 | 96 |
| $A\bar{B}C$ | 32 | 128 | 128 | 512 |
| $AB\bar{C}$ | 1024 | 1536 | 4096 | 6144 |
| $ABC$ | 128 | 512 | 512 | 2048 |

Table 3: Statistical information in relation ABCDE

Thus, we can use Algorithm 2.1 to decompose ABCDE into 4NF subrelations AB, AC, BCD, CE. Each relation preserves the corresponding dependency and they have a lossless join equal to ABCDE. The statistical information in subrelations is shown in Table 4.

|  | $\bar{A}$ | $A$ |
|---|---|---|
| $\bar{B}$ | 64000 | 4000 |
| $B$ | 16000 | 16000 |

(a)

|  | $\bar{A}$ | $A$ |
|---|---|---|
| $\bar{C}$ | 7600 | 16000 |
| $C$ | 4000 | 4000 |

(b)

|  | $\bar{B}\bar{C}$ | $\bar{B}C$ | $B\bar{C}$ | $BC$ |
|---|---|---|---|---|
| $\bar{D}$ | 60800 | 800 | 5600 | 800 |
| $D$ | 3200 | 3200 | 22400 | 3200 |

(c)

|  | $\bar{C}$ | $C$ |
|---|---|---|
| $\bar{E}$ | 36800 | 1600 |
| $E$ | 55200 | 6400 |

(d)

Table 4: Statistical information in sub-relations

# 3   BELIEF NETWORKS

In this section, we introduce some concepts of BN based on the theory of discrete Markov random fields.

## 3.1   PROBABILISTIC DEPENDENCY

Data extracted from a RD are called relational data. Although statistical relational data may satisfy MD, it refers to only FD and probabilistic dependency, a special kind of MD.

**Definition 3.1:** The **frequency** of an attribute subset $X$ of relation scheme $R$ in a relation $r$ on $R$ is

$F_X(r) = \{F_{X=x}(r) = \frac{|\sigma_{X=x}(r)|}{|r|}; \forall x \in D_X\}$ where $D_X$ is the domain of $X$, $|r|$ is the cardinality of $r$.

**Definition 3.2:** The **conditional frequency** of $X$ in $r$ given $Y = y$, $Y \subset R$ is $F_{X|Y=y}$:

$$\{F_{X=x|Y=y}(r) = \frac{|\sigma_{X=x \wedge Y=y}(r)|}{|\sigma_{Y=y}(r)|}; \forall x \in D_X\}$$

**Definition 3.3:** Let $X, Y \subset R$, and $Z = R - XY$. $r$ satisfies the **probabilistic dependency** (PD) $X \mapsto Y$ if $xyz \in r \Longrightarrow$

$$F_{XYZ=xyz}(r) \cdot F_{X=x}(r) = F_{XY=xy}(r) \cdot F_{XZ=xz}(r).$$

According to the law of large numbers, it is reasonable to assume $\lim_{|r| \to \infty} F_X(r) = P(X)$, $\lim_{|r| \to \infty} F_{X|Y=y}(r) = P(X|Y=y)$, and if $X \mapsto Y$ then $P(Y|XZ) = P(Y|X)$, ie. $Y$ and $Z$ are conditionally independent given $X$. It is easy to prove

**Theorem 3.1:**

1. $X \to Y \Longrightarrow X \mapsto Y$ with $P(Y|X) \in \{0, 1\}$, and

2. $X \mapsto Y \Longrightarrow X \longrightarrow\!\!\!\rightarrow Y$.

In Example 2, we can easily check that the PD $A \mapsto B$, $A \mapsto C$, $B, C \mapsto D$, and $C \mapsto E$ hold.

## 3.2   BASIC DEFINITIONS OF BELIEF NETWORKS

Consider a probability space $R = \{A_i | i = 1, ...m\}$, which corresponds to a relation scheme $R$, with $M = a_1 \times ... \times a_m$ possible states $S = \{s_j | j = 1, ..., M\}$ and a probability distribution $p = \{P(s_j) | j = 1, ..., M\}$. Each variable $A_i$ in the space can take $a_i$ values. Suppose according to the dependencies, we have the following constraint set $CS$ on the distribution $p$ of $R$:

**Conditional constraints (CCS):**

$$\mu_k = P(A_{k0} | X_{k1, ..., kp_k})$$

corresponding to the PD $X_{k1, ..., kp_k} \mapsto A_{k0}$, where $k = 1, ..., n$ and $X_{k1, ..., kp_k} = \{A_{k1}, ..., A_{kp_k}\} \subset R$.

Marginal constraints ($MCS$):

$$\nu_{k'} = P(X_{k'1, ..., k'p'_{k'}}), \ where \ k' = 1, ... n'.$$

Universal constraint (UCS):

$$\sum_{A_0, ..., A_{m-1}} P(A_0, ..., A_{m-1}) = 1.$$

In Example 1, it is easy to extract the CCS in Table 5 from the sample database in Table 2. In Table 5, $u$ stands for $u = 1$, $\bar{u}$ for $u = -1$ and $\hat{u}$ for $u = 0$. Note that all these conditional probabilities are accurate and equal to 1, because all of the dependencies

$$\begin{array}{llll}
P(\bar{u}_7|\bar{u}_1,\bar{u}_2,\bar{u}_3) & =1, & P(u_7|\bar{u}_1,\bar{u}_2,u_3) & =1 \\
P(\bar{u}_7|\bar{u}_1,u_2,u_3) & =1, & P(u_7|u_1,\bar{u}_2,\bar{u}_3) & =1 \\
P(u_7|\bar{u}_1,\bar{u}_2,\bar{u}_3) & =1, & P(\bar{u}_7|u_1,u_2,\bar{u}_3) & =1 \\
P(u_7|u_1,u_2,u_3) & =1, & & \\
\end{array}$$

$$\begin{array}{llll}
P(\bar{u}_8|\bar{u}_3,\bar{u}_4,u_5) & =1, & P(u_8|\bar{u}_3,u_4,u_5) & =1 \\
P(u_8|\bar{u}_3,u_4,u_5) & =1, & P(\bar{u}_8|u_3,\bar{u}_4,\bar{u}_5) & =1 \\
P(\bar{u}_8|u_3,\bar{u}_4,\bar{u}_5) & =1, & P(\bar{u}_8|\bar{u}_3,\bar{u}_4,\bar{u}_5) & =1 \\
P(u_8|u_3,u_4,\bar{u}_5) & =1, & & \\
\end{array}$$

$$\begin{array}{llll}
P(\bar{u}_9|\bar{u}_6,\bar{u}_7,\bar{u}_8) & =1, & P(u_9|\bar{u}_6,\bar{u}_7,u_8) & =1 \\
P(u_9|\bar{u}_6,u_7,\bar{u}_8) & =1, & P(u_9|\bar{u}_6,u_7,u_8) & =1 \\
P(\bar{u}_9|u_6,\bar{u}_7,u_8) & =1, & P(\bar{u}_9|u_6,u_7,\bar{u}_8) & =1 \\
P(\bar{u}_9|u_6,u_7,u_8) & =1, & & \\
\end{array}$$

$$\begin{array}{llll}
P(\bar{u}_{10}|\bar{u}_3,\bar{u}_7,\bar{u}_8) & =1, & P(u_{10}|\bar{u}_3,\bar{u}_7,u_8) & =1 \\
P(u_{10}|\bar{u}_3,u_7,u_8) & =1, & P(u_{10}|\bar{u}_3,u_7,\bar{u}_8) & =1 \\
P(\bar{u}_{10}|u_3,\bar{u}_7,u_8) & =1, & P(\bar{u}_{10}|u_3,u_7,\bar{u}_8) & =1 \\
P(\bar{u}_{10}|u_3,u_7,u_8) & =1, & & \\
\end{array}$$

$$\begin{array}{llll}
P(\bar{u}_{11}|\bar{u}_9,\bar{u}_{10}) & =1, & P(u_{11}|\bar{u}_9,u_{10}) & =1 \\
P(u_{11}|u_9,\bar{u}_{10}) & =1, & P(\bar{u}_{11}|u_9,u_{10}) & =1 \\
\end{array}$$

Table 5: The conditional constraint sets from Table 2

are FD. It may not be the case for PD, ie. the conditional probabilities may not be accurately estimated nor they necessarily equal to 1. Note also that the above set of conditional probabilities is not complete, eg. $P(u_7|\bar{u}_1,u_2,\bar{u}_3)$ is not specified.

According to data dependencies and CCS, we may construct a directed graph, or BN as follows

**Definition 3.4:** A **Belief Network (BN)** is a directed graph $G = \langle V, E \rangle$, such that

1. The node set is $V = R$

2. The edge set is $E = \{\langle A_{kq}, A_{k0} \rangle\}$, such that

$$\exists CCS \; \mu_k = P(A_{k0}|X_{k1,...,kp}) \in CS, A_{kq} \in X_{k1,...,kp}.$$

In Example 1, we have a BN shown in Fig. 1.

**Definition 3.5:**

A **neighbor system** $\sigma$ in $G$ is a set of sets $\{\sigma A_i | A_i \in R, \sigma A_i \subseteq R\}$, such that

1. $A_i \notin \sigma A_i$,

2. $A_j \in \sigma A_i \iff \exists \; CCS \; \mu_k = P(A_{k0}|X_{k1,...,kp}) \in CS, A_i, \; A_j \in X_{k0,...,kp}$

The **neighbors of a set** $X \subset R$ in G is the set $\sigma X = \{A_i \in R - X | \exists A_j \in X, A_i \in \sigma A_j\}$.

The **neighborhood network** of a BN $G = \langle V, E \rangle$ is $G_\sigma = \langle V, E_\sigma \rangle$, where $E_\sigma = \{(A_i, A_j) | A_i \in \sigma A_j\}$.

A set $C \subseteq R$ is called a **clique** if $A_j \in \sigma A_i$ whenever $A_i, A_j \in C$ and $i \neq j$. A clique $MC$ is called **maximal**

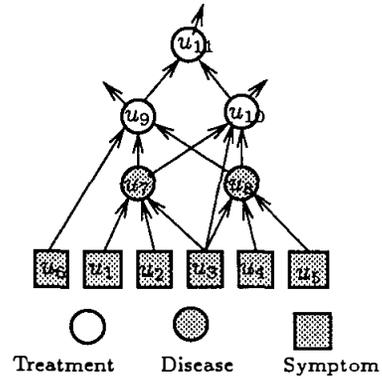

Figure 1: Belief Network of Sarcophagal disease

**clique** if there is no other clique $C$, such that $MC \subset C$. Let $MCC$ be the class of all maximal cliques in $R$.

For Example 1, the neighborhood network is given in Fig. 2.

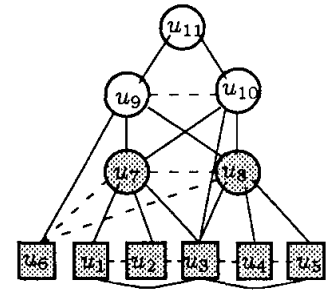

Figure 2: The neighborhood network for Example 2

### 3.3 DECOMPOSITION OF BELIEF NETWORKS

In order to handle the combinatorial explosion of the number of states in BN, a decomposition may be desired. The concept of neighbor Gibbs field [Wen, 1989] provides a valid factorization of the joint distribution for Markov random fields to localize the computation of the joint ME/MCE distribution of the whole BN within each of the maximal cliques of the neighborhood network. This suggests that the network should be decomposed into a hypergraph with the cliques of the neighborhood network as its hyperedges. To keep consistency among the distributions of the cliques, the results obtained in each clique need to be propagated to other cliques through their intersections. Consequently, it is desired to organize the decomposed result as an acyclic hypergraph [Beeri *et al.*, 1983] to guarantee the termination of the propagation and to avoid other possible anomalies during the propagation.

The decomposition techniques proposed in [Spiegelhalter, 1986; Wen, 1991] are described briefly as follows:

1. Construct a neighborhood network $G_\sigma = <V, E_\sigma>$ for BN $G = <V, E>$.

2. Find a fill-in [Tarjan and Yannakakis, 1984] $F$ of $G_\sigma$, such that $D_\sigma$, the $MCC$ of $G_f = <V, F \cup E_\sigma>$

   • has the minimum $|F|$, for Spiegelhalter's method,

   • has the minimum total number of states of all cliques in $G_f$ for our method [Wen, 1991].

$D_\sigma$ is the decomposition wanted and corresponds to an acyclic hypergraph $<V, D_\sigma>$.

Unfortunately, it has been shown that the problem of optimum belief network decomposition is NP hard under all of the above optimum criteria [Yannakakis, 1981; Wen, 1991]. Therefore, we proposed an algorithm to obtain the optimum belief decomposition based on simulated annealing [Wen, 1991].

In Example 1, it is easy to verify by Graham reduction [Beeri et al., 1983] that the neighborhood network in Fig. 2 has already been an acyclic hypergraph. The decomposition is shown in Fig. 3. There are 6 sub-netwroks in the decomposition: $MC_1 = \{u_9, u_{10}, u_{11}\}$, $MC_2 = \{u_7, u_8, u_9, u_{10}\}$  $MC_3 = \{u_6, u_7, u_8, u_9\}$, $MC_4 = \{u_3, u_7, u_8, u_{10}\}$, $MC_5 = \{u_3, u_4, u_5, u_8\}$, and $MC_6 = \{u_1, u_2, u_3, u_7\}$. The total number of states here is 124, comparing $2^9 \times 3^2 = 4608$ states in the original BN.

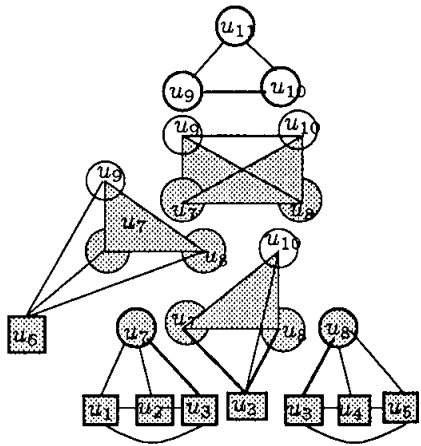

Figure 3: Decomposition of the BN for Example 2

Note that the decomposition corresponding to an acyclic database scheme [Beeri et al., 1983] in Table 6, each relation scheme corresponds to a maximal clique in the decomposition. These relation schemes also satisfy JD and thus have lossless join and running intersection property [Beeri et al., 1983]. In Fig. 3, the intersections between cliques have been shown by bold edges or shaded triangles.

| $u_1$ | $u_2$ | $u_3$ | $u_7$ | $u_3$ | $u_4$ | $u_5$ | $u_8$ |
|---|---|---|---|---|---|---|---|
| 1 | 1 | 1 | 1 | 1 | -1 | 0 | -1 |
| -1 | -1 | -1 | -1 | -1 | 1 | 1 | 1 |
| -1 | -1 | 1 | 1 | 1 | 1 | -1 | 1 |
| 1 | 1 | -1 | -1 | -1 | -1 | 1 | -1 |
| 1 | -1 | 0 | 1 | 0 | 1 | 1 | 1 |
| 1 | -1 | -1 | 1 | -1 | 1 | 1 | 1 |
| 1 | 1 | 1 | 1 | 1 | -1 | -1 | 1 |
| -1 | 1 | 1 | -1 | 1 | -1 | 1 | 1 |

| $u_6$ | $u_7$ | $u_8$ | $u_9$ | $u_3$ | $u_7$ | $u_8$ | $u_{10}$ |
|---|---|---|---|---|---|---|---|
| -1 | 1 | -1 | 1 | 1 | 1 | -1 | 1 |
| -1 | -1 | 1 | 1 | -1 | -1 | 1 | -1 |
| 1 | 1 | 1 | -1 | 1 | 1 | 1 | -1 |
| -1 | -1 | -1 | -1 | -1 | -1 | -1 | -1 |
| 1 | 1 | 1 | -1 | 0 | 1 | 1 | 1 |
| -1 | 1 | 1 | 1 | -1 | 1 | 1 | 1 |
| 1 | 1 | -1 | -1 | 1 | 1 | -1 | 1 |
| 1 | -1 | 1 | -1 | 1 | -1 | 1 | 1 |

| $u_9$ | $u_{10}$ | $u_{11}$ | $u_7$ | $u_8$ | $u_9$ | $u_{10}$ |
|---|---|---|---|---|---|---|
| 1 | -1 | 1 | 1 | -1 | 1 | -1 |
| 1 | 1 | -1 | -1 | 1 | 1 | 1 |
| -1 | 1 | -1 | 1 | 1 | -1 | 1 |
| -1 | -1 | -1 | -1 | -1 | 1 | -1 |
| -1 | 1 | 1 | 1 | 1 | -1 | 1 |
| 1 | -1 | -1 | 1 | 1 | 1 | 1 |
| -1 | -1 | -1 | 1 | -1 | -1 | -1 |
| -1 | -1 | -1 | -1 | -1 | -1 | -1 |

Table 6: An acyclic database for Table 2

In [Wen, 1989], we have proven

**Theorem 3.2:** Belief updating the whole BN with Jeffrey's rule [Wen, 1990b] is equivalent to belief updating the clique in the acyclic decomposition which contains the corresponding constraint set, and Jeffrey belief propagation to all the other cliques through the running intersections.

## 3.4 THE MAIN OPERATIONS ON BN

There are three main operations on BN:

**Belief extracting** – Extracting a specified marginal distribution of a distribution. This corresponds to the projection operation in RD.

**Updating** – Given a new marginal on a subspace and a prior on the whole space, calculate a plausible posterior of the whole space matching with the given marginal. Bayes or Minimum Cross Entropy (MCE) posterior, particularly the posterior obtained by Jeffrey's updating [Wen, 1990b], are considered as plausible. This corresponds to selection operation for RD.

**Belief propagation** – After updating the marginal of a subspace, propagate the changes to the whole space through the running intersections. This operation corresponds to the join operation in RD.

## 4  INITIAL DISTRIBUTIONS

For the initial distribution of a decomposed network, two requirements should be satisfied.

1. The distribution should reflect the data in the sample database as faithfully as possible.

2. The distribution should predicate unseen cases as accurately as possible.

### 4.1  RECALL

There may be many distributions satisfying the fist requirement if the specification is incomplete. The most trivial one can be constructed as follows:

In Table 2, use 0 and 1 to replace of -1 and 1 for binary variables and use 00, 01 and 10 to replace of -1, 0 and 1 for $u_3$ and $u_5$, and convert each row in each subrelation into hexadecimal, then we obtain "Index" in Table 7. Suppose all these examples are equally important, or have the same probability, we obtain "Probabilities" in Table 7.

| $MC_1$ | | $MC_2$ | | $MC_3$ | |
|---|---|---|---|---|---|
| Index: | Freq. | Index: | Freq. | Index: | Freq. |
| 5: | 0.125 | 0: | 0.125 | 5: | 0.125 |
| 6: | 0.25 | 4: | 0.125 | 3: | 0.125 |
| 0: | 0.5 | 7: | 0.125 | e: | 0.25 |
| 3: | 0.125 | 8: | 0.125 | 0: | 0.125 |
| | | a: | 0.125 | 7: | 0.125 |
| | | c: | 0.125 | a: | 0.125 |
| | | d: | 0.125 | c: | 0.125 |
| | | f: | 0.125 | | |

| $MC_4$ | | $MC_5$ | | $MC_6$ | |
|---|---|---|---|---|---|
| Index: | Freq. | Index: | Freq. | Index: | Freq. |
| 14: | 0.25 | 22: | 0.125 | 1d: | 0.25 |
| 03: | 0.125 | 0d: | 0.25 | 00: | 0.125 |
| 16: | 0.125 | 29: | 0.125 | 05: | 0.125 |
| 00: | 0.125 | 04: | 0.125 | 18: | 0.125 |
| 0f: | 0.125 | 1d: | 0.125 | 13: | 0.125 |
| 07: | 0.125 | 20: | 0.125 | 11: | 0.125 |
| 12: | 0.125 | 25: | 0.125 | 0c: | 0.125 |

Table 7: Prior distributions for cliques

Thus, Table 7 gives all the non-zero probabilities in a distribution satisfying all of the training examples. Using this simple distribution, we can perform all recall-like reasoning. Suppose we are given

$u_1 = 1$: a patient has swollen feet,
$u_3 = 1$: the patient suffers from hair loss,
$u_6 = -1$: the patient is not allergic to placibin.

This corresponds to a constraint set $\{P(u_1) = 1, P(u_3) = 1, P(u_6) = 0\}$. By Jeffrey's updating and belief propagation, we obtain a posterior in Table 8. This implies

(1) $u_2 = u_7 = u_9 = u_{11} = 1$,  (2) $u_5 = 0$,
(3) $u_4 = u_8 = u_{10} = -1$.

Comparing with the result in [Gallant, Oct 1987]:

| | |
|---|---|
| $MC_1$ : $P(u_9 = 1, u_{10} = -1, u_{11} = 1)$ | $= 1$ |
| $MC_2$ : $P(u_7 = 1, u_8 = -1, u_9 = 1, u_{10} = -1)$ | $= 1$ |
| $MC_3$ : $P(u_6 = -1, u_7 = 1, u_8 = -1, u_9 = 1)$ | $= 1$ |
| $MC_4$ : $P(u_3 = 1, u_7 = 1, u_8 = -1, u_{10} = -1)$ | $= 1$ |
| $MC_5$ : $P(u_3 = 1, u_4 = -1, u_5 = 0, u_8 = -1)$ | $= 1$ |
| $MC_6$ : $P(u_1 = 1, u_2 = 1, u_3 = 1, u_7 = 1)$ | $= 1$ |

Table 8: Posterior distributions for cliques

(1) $u_7 = u_9 = u_{11} = 1$,  (2) $u_{10} = -1$,

we can see that our method do reasoning in all directions while Gallants can only do reasoning bottom-up.

This method is simple but has some disadvantages:

1. There are still "unseen" cases that cannot be handled by this method.

2. Not all MCS extracted from an RD are always consistent.

To overcome the second difficulty, we should extract a set of conditional probabilities, or *Local Characteristics (LC)*, instead of marginal ones, because any distribution is completely determined by its LC's and a set of non-redundant LC's can be always made consistent.

### 4.2  PREDICTION OR GENERALIZATION

It is obviously more difficult to predicate unseen cases than to just recall the cases encountered before.

Having constructed a BN, we can use one of the following methods to learn a set of LC's from an incomplete training database:

**Frequency Method:** A set of LC's can be learned in the same way as that extracting marginal probabilities described in the previous subsection. The disadvantage is that it has no generalization ability at all.

**ME/MCE methods** assign a uniform conditional distribution to the unseen cases. A special case of ME/MCE methods is so called Dirichlet distribution [Herskovitz and Cooper, 1990] which uses the following formula

$$P(X = x | \Pi_X = \pi_X) = \frac{C(X = x, \Pi_X = \pi_X) + 1}{C(\Pi_X = \pi_X) + V_X}$$

where $X$ is a variable in the underlying BN, $x$ is one of the $V_X$ values can be taken by $X$, $\Pi_X$ is the set of parents of $X$ in the BN, $\pi_X$ is a particular instantiation of $\Pi_X$, and $C(\Phi)$ is the number of cases/tuples in the database that match the instantiated set of variables $\Phi$. When the case/tuple does not occur in the database, the above conditional probability becomes $P(X = x | \Pi_X = \pi_X) = \frac{1}{V_X}$ and thus is a uniform one.

**The NN/OR Method** [Duda and Hart, 1973; Blumer *et al.*, 1987]: For this method, the conditional probability assigned to an unseen case depends on its neighbor conditional distributions. That is, if the unseen case has many neighbors who have high probabilities to occur then it is assigned a relatively high conditional probability, otherwise, a low or even zero conditional probability. When there are neighbors having different probabilities, the NN/OR method prefers the choice making the final result simplest.

### 4.3 THE NN/OR LEARNING

It has been shown that the decomposed relations preserve all dependencies existing in the original relation. For Example 1, this means that learning can be performed within each relation (see Table 2). For example, in relation $\{u_1, u_2, \bar{u}_3, u_7\}$, conditional distribution $P(u_7|u_1, u_2, u_3)$ can be learned as follows:

1. Use frequency method to obtain the conditional probabilities for the cases occurring in the training database. For relation $(u_1, u_2, u_3, u_7)$ in Table 2. The result is

$$
\begin{array}{llll}
P(u_7|\bar{u}_1, \bar{u}_2, \bar{u}_3) & = 0 & P(u_7|\bar{u}_1, \bar{u}_2, u_3) & = 1 \\
P(u_7|\bar{u}_1, u_2, u_3) & = 0 & P(u_7|u_1, \bar{u}_2, \bar{u}_3) & = 1 \\
P(u_7|u_1, u_2, \bar{u}_3) & = 0 & P(u_7|u_1, u_2, u_3) & = 1
\end{array}
$$

2. Draw a Karnough-like map for $u_1$, $u_2$, $u_3$ and fill the probability values learned in step 1 into the corresponding entries of the map (see Fig.4 a).

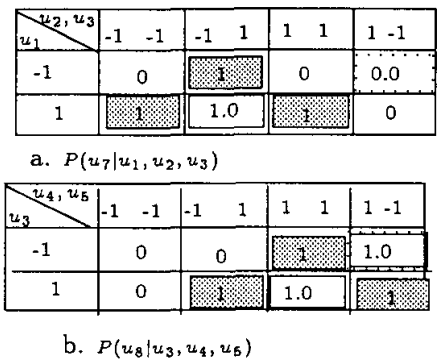

a. $P(u_7|u_1, u_2, u_3)$

b. $P(u_8|u_3, u_4, u_5)$

Figure 4: The results of learning by NN/OR method

3. For the unseen cases, entry (-1,1,-1) is filled with value 0 because all its nearest neighbors have entry values 0, and entry (1,-1,1) is filled with 1, similarly.

In some cases, the entries in the nearest neighborhood have different values. For example, $P(u_8|\bar{u}_3, u_4, \bar{u}_5)$ (see Fig. 4 b) has two nearest neighbors, (-1,1,1) and (1,1,-1), with values 1 and another one, (-1,-1,-1) with value 0. In this case, the Occam's Razor principle [Blumer *et al.*, 1987] can be used to choose the value for the unseen cases. The principle says

Among the hypotheses consistent or compat-ible with the given data set, choose the simplest one.

There are many proposed measures of simplicity, the most common ones are as follows

1. Kolmogorov complexity,

2. Minimum description length, and

3. Logic formula complexity, combinational complexity, and time complexity [Pearl, 1978].

We adopt logic formula complexity which depends on the number of connectives in the logic formula. Trying to assign 0 and 1 to entry $(\bar{u}_3, u_4, \bar{u}_5)$, respectively, we find that logic formula $u_4 + u_3 u_5$ corresponding to $P(u_7|\bar{u}_3, u_4, \bar{u}_5) = 1$ is simpler than $u_3 u_4 + u_3 u_5 + u_4 u_5$ corresponding to $P(u_7|\bar{u}_3, u_4, \bar{u}_5) = 0$. Therefore, we choose $P(u_7|\bar{u}_3, u_4, \bar{u}_5) = 1$.

The logical expressions of the 4 possible assignments for unseen cases in Fig. 4 a. are give in Table 9. Obviously, the one obtained by NN/OR method is the simplest and has the shortest code length.

| $p_2$ | $p_5$ | Logic expression | Complex. |
|---|---|---|---|
| 0 | 0 | $\bar{u}_1 \bar{u}_2 u_3 + u_1 \bar{u}_2 \bar{u}_3 + u_1 u_2 u_3$ | 8 |
| 0 | 1 | $u_1 \bar{u}_2 + u_1 u_3 + \bar{u}_2 u_3$ | 5 |
| 1 | 0 | $u_1 \bar{u}_2 u_3 + u_1 \bar{u}_2 \bar{u}_3 + u_1 u_2 u_3 + \bar{u}_1 u_2 \bar{u}_2$ | 11 |
| 1 | 1 | $u_1 \bar{u}_2 + u_1 u_3 + \bar{u}_2 u_3 + \bar{u}_1 u_2 \bar{u}_2$ | 8 |

$p_2 = P(u_7|\bar{u}_1, u_2, \bar{u}_3)$,    $p_5 = P(u_7|u_1, \bar{u}_2, u_3)$

Table 9: Assignments for unseen cases in Fig. 3 a

Fig. 5 gives the results when some statistical information in Table 4 c is missing.

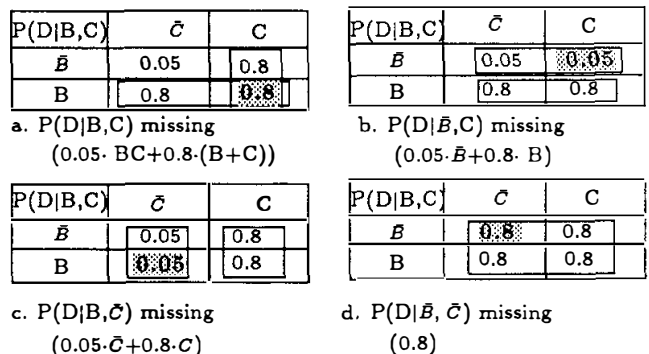

a. P(D|B,C) missing
(0.05· BC+0.8·(B+C))

b. P(D|$\bar{B}$,C) missing
(0.05·$\bar{B}$+0.8· B)

c. P(D|B,$\bar{C}$) missing
(0.05·$\bar{C}$+0.8·$C$)

d. P(D|$\bar{B}$, $\bar{C}$) missing
(0.8)

Figure 5: Learning $P(D|B, C)$ in Example 2

Note that we are using these examples to describe the method proposed here, we are not saying that any required learning accuracy can be guaranteed by the given sample sets. For more detail about learning from statistical relational data, see [Wen, 1990a].

## 5  CONCLUSIONS

The relationship between RD and BN is investigated. Correspondences are discovered between many con-cepts and operations of RD and BN. A method to construct BN automatically from RD is proposed. It has been shown [Wen, 1990a] that the distributions of discrete Markov fields, eg. BN, are Probably Approximately Correctly (PAC) learnable [Haussler, 1990] when the sizes of the biggest neighborhoods of the variables in the fields are fixed. Hence, our method, is efficient in these circumstances. A comparison between our method and other methods shows that

1. Our method can fulfill the task of recall perfectly just as some other methods.

2. Our method has more plausible result than that of other methods when generalization or prediction is needed. For example, frequency method does not have any prediction capability while ME/MCE methods can not handle the case of functional dependency properly (always assign values $\frac{1}{2}$ to all binary unseen cases).

## Acknowledgement

Thanks to A. Jennings and H. Liu for discussions and comments. The permission of the Executive General Manager, TRL, to publish this paper is gratefully acknowledged.